\documentclass[letterpaper, 10 pt, conference]{ieeeconf}  

\IEEEoverridecommandlockouts                              

\usepackage{graphicx} 
\usepackage[table]{xcolor}
\usepackage{adjustbox}
\usepackage{booktabs}
\usepackage{caption}
\usepackage{mathptmx} 
\usepackage{amsmath} 
\usepackage{amssymb}  
\usepackage{bm}
\usepackage{lipsum}
\usepackage{subcaption}
\usepackage{cite}

\DeclareMathOperator*{\argmax}{arg\,max}

\title{\LARGE \bf
Multimodal Anomaly Detection with a Mixture-of-Experts
}

\author{Christoph Willibald$^{ \star, 1}$ and Daniel Sliwowski$^{\star, 2}$ and Dongheui Lee$^{1,2}$
\thanks{$\star$ Equal contribution.}
\thanks{$^{1}$Institute of Robotics and Mechatronics (DLR), German Aerospace Center, Wessling, Germany.}%
\thanks{$^{2}$Autonomous Systems Lab, Institute of Computer Technology, TU Wien, Vienna, Austria.}%
\thanks{This work has been partially supported by the European Union project INVERSE under grant agreement No. 101136067, and in part by the Robot Industry Core Technology Development Program under Grant 00416440 funded by the Korea Ministry of Trade, Industry and Energy (MOTIE).}
}

\begin{document}

\maketitle
\thispagestyle{empty}
\pagestyle{empty}

\begin{abstract}
With a growing number of robots being deployed across diverse applications, robust multimodal anomaly detection becomes increasingly important. In robotic manipulation, failures typically arise from (1) robot-driven anomalies due to an insufficient task model or hardware limitations, and (2) environment-driven anomalies caused by dynamic environmental changes or external interferences. Conventional anomaly detection methods focus either on the first by low-level statistical modeling of proprioceptive signals or the second by deep learning-based visual environment observation, each with different computational and training data requirements. To effectively capture anomalies from both sources, we propose a mixture-of-experts framework that integrates the complementary detection mechanisms with a visual-language model for environment monitoring and a Gaussian-mixture regression-based detector for tracking deviations in interaction forces and robot motions. We introduce a confidence-based fusion mechanism that dynamically selects the most reliable detector for each situation. We evaluate our approach on both household and industrial tasks using two robotic systems, demonstrating a 60\% reduction in detection delay while improving frame-wise anomaly detection performance compared to individual detectors.

\end{abstract}

\section{Introduction}

\begin{figure}
    \centering
    \includegraphics[width=\linewidth]{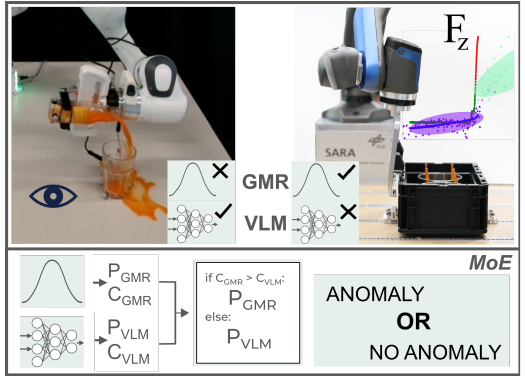}
    \caption{Our Mixture of Experts (MoE) framework that combines GMR-based outlier detection with VLM-based scene monitoring to identify anomalies in both household and industrial settings. The framework dynamically selects the most reliable detector for each situation based on a prediction confidence score.}
    \label{fig:Teaser}
\end{figure}

As collaborative robots act increasingly autonomously across various applications, accurately monitoring task progress and success becomes crucial. In both household and industrial settings, like the ones depicted in Fig.~\ref{fig:Teaser}, autonomous robots are confronted with a range of unknown situations, leading to diverse sources of task failure. Common anomalies arise from hardware defects, human interference, or inaccuracies in the task model. To account for those diverse failure cases, an anomaly detection mechanism must integrate multimodal sensor data to decide whether the current situation constitutes an anomaly.

Conventional anomaly detectors in robotics, whether online or offline, typically follow two approaches: modeling the statistical distribution of low-dimensional data to identify outliers or leveraging deep neural networks to detect anomalies from high-dimensional inputs like video~\cite{10.1145/1541880.1541882}. Approaches of the former category \cite{willibald2020collaborative, eiband2023collaborative, Eiband2019, romeres2019anomaly, park2019multimodal, azzalini2020hmms} perform well in unsupervised detection of deviations from proprioceptive measurements with sparse training data. For that, probabilistic distributions of expected measurements are learned from successful executions to detect deviations such as anomalous process forces with different thresholding methods. Deep learning-based detectors ~\cite{sliwowski2024conditionnet, park2018multimodal, AnomalyDetSlip, FaluireClassification1, Altan2022CLUEAIAC, chalapathy2017robust, SuccessVQA, FinoNet, TPVQA, Sinha-RSS-24} utilize a supervised of self-supervised setting, typically requiring large training datasets of both positive and negative task examples, which makes them more effective at recognizing failures in environmental interactions, such as spilling liquid during pouring. Since the different detection methods vary in their training data requirements and computational complexity, it is beneficial to train individual detectors on distinct data subsets. However, to robustly detect anomalies in robotic manipulation tasks, a combined identification of deviations in motion, force, and interaction dynamics between the robot and its environment is required. Therefore, a comprehensive anomaly detection system must integrate multiple specialized subdetectors into a unified framework.

We propose a novel multimodal anomaly detection strategy for robust task monitoring, integrating multiple detection approaches within a mixture-of-experts framework. Our method combines two complementary detectors: one analyzing the robot’s low-level dynamics and the other visually monitoring the scene to detect deviations from the expected task execution. A visual-language model (VLM) \cite{sliwowski2024conditionnet} detects violations of action preconditions and effects resulting from interaction with the environment, while a Gaussian-mixture regression (GMR)-based detector~\cite{willibald2020collaborative} continuously tracks deviations in interaction forces and robot poses. Considering the prediction confidence of each approach, their outputs are fused to produce a joint result. This late fusion approach allows the framework to dynamically select the most suitable detector in a given situation. Figure~\ref{fig:Teaser} highlights scenarios where either VLM-based or GMR-based anomaly detection is most effective, emphasizing the versatility and advantages of a fused approach.

The contributions of our work are as follows:
\begin{enumerate}
    \item a visual-language anomaly detection approach based on \cite{sliwowski2024conditionnet} which uses task progress to ground the expected action state and obtain a prediction confidence score;
    \item a probabilistic anomaly detection method based on~\cite{willibald2020collaborative} using local modality-specific anomaly thresholds and quantification of prediction confidence;
    \item a mixture-of-experts fusion strategy for merging probabilistic and deep-learning-based anomaly detection approaches.
\end{enumerate}

\section{Related works}
Previous works in robotics attempt to solve the anomaly detection problem by using either probabilistic modeling or deep learning.
\subsection{Probabilistic anomaly detection}
Probabilistic approaches in ~\cite{willibald2020collaborative, Eiband2019, eiband2023collaborative} model successful task executions with Gaussian Mixture Models (GMM). By regressing the model on time, these methods compute expected sensor measurements and dynamically adjust anomaly detection sensitivity through probabilistic modeling. Chernova et al. \cite{Chernova2007} employ a GMM to encode a simple policy using basic symbolic actions and utilize the observation likelihood of unseen states for outlier detection. An offline anomaly detection method is presented in~\cite{romeres2019anomaly}, where Gaussian Processes Regression is used to predict force profiles and epistemic uncertainty during insertion tasks to compute an anomaly score at every time step. A run is unsuccessful if the average anomaly score over all time steps exceeds a predefined threshold. Probabilistic detection of anomalies can also be achieved with a Hidden Markov Model, either through predefined anomaly thresholds combined with a sliding time window approach \cite{azzalini2020hmms}, or by estimating probabilistic thresholds based on execution progress \cite{park2019multimodal}. These methods treat anomaly detection as statistical outlier detection of low-dimensional features derived from multimodal sensor data. However, unlike our VLM-based detector, they do not incorporate visual observations of the scene to identify anomalous effects of the robot’s actions on the environment.

In \cite{agia2024unpacking}, the action consistency of a diffusion policy is evaluated across consecutive timesteps using a statistical distance function. An anomaly is detected when the cumulative sum of these distances exceeds a threshold learned from successful executions. This detector is combined with an LVLM to assess task progress through chain-of-thought reasoning and video question-answering, determining whether the robot is still progressing toward the task goal. However, unlike our approach, this method cannot detect deviations in contact force and exhibits significantly longer response times for the LVLM-based detector.

\subsection{Deep-Learning anomaly detection}
Probabilistic approaches perform well with low-dimensional data, such as end-effector poses or force-torque measurements, but struggle with high-dimensional data like images or audio~\cite{10.1145/1541880.1541882}. Deep-learning methods address this by learning feature representations while detecting anomalies~\cite{park2018multimodal, AnomalyDetSlip, FaluireClassification1, Altan2022CLUEAIAC}. Broadly, deep-learning anomaly detection techniques fall into four categories: reconstruction-based approaches, one-class neural networks, success detectors, and variants of visual- and large language models.  

Reconstruction-based methods treat anomaly detection as a self-supervised task, training generative models such as autoencoders~\cite{AD_AE}, variational autoencoders~\cite{park2018multimodal}, or generative adversarial networks~\cite{GAN_AD} to encode and decode observations. Since they are trained only on ``normal'' data, they struggle to reconstruct anomalies, which results in a higher reconstruction loss. Anomalies are detected by thresholding this loss. These methods have been successfully applied in robotics tasks, such as autonomous feeding~\cite{park2018multimodal}.  
One-class neural networks extend the one-class support vector machine to deep architectures~\cite{chalapathy2017robust}. They introduce loss functions that separate normal data representations using a hyperplane or hypersphere in latent space. Instead of reconstruction loss, they produce an ``outlier score'' that is thresholded to detect anomalies.  

Success detection approaches are specific to robotic task execution anomaly detection, framing anomaly detection as a supervised classification problem. A neural network classifies executions as successful or unsuccessful based on single or sequential observations from the execution. These methods require both positive and negative task demonstrations. Vision-language models like SuccessVQA~\cite{SuccessVQA} apply this concept, while FinoNET~\cite{FinoNet} integrates video and audio data to make the predictions.

Lastly, ``other'' approaches aggregate deep learning methods combining vision and natural language descriptions. Some of these methods leverage large language models (LLMs) and large vision-language models (LVLMs) for reasoning-based anomaly detection. TP-VQA~\cite{TPVQA} queries an LVLM by asking whether predicates in a task's planning domain definition language (PDDL)~\cite{PDDL} description are satisfied. AESOP~\cite{Sinha-RSS-24} detects anomalies by comparing text embeddings of current and ``normal'' scene descriptions, using an LLM to confirm significant differences. Alternatively, ConditionNET~\cite{sliwowski2024conditionnet} learns action preconditions and effects and detects anomalies by comparing the predicted and expected action states.  

Deep-learning approaches also allow leveraging of multimodal information, by appropriately designing the network architecture. Works like~\cite{AnomalyDetSlip,  FaluireClassification1, Altan2022CLUEAIAC} adopt a late fusion strategy, where features from each modality are extracted separately and then concatenated and classified with a Multi-Layer perception. Alternatively, works like~\cite{park2018multimodal} adopt an early fusion strategy, where raw sensory readings are concatenated, and the model learns to reconstruct the data for successful executions.

\begin{figure*}
    \centering
    \includegraphics[width=\linewidth]{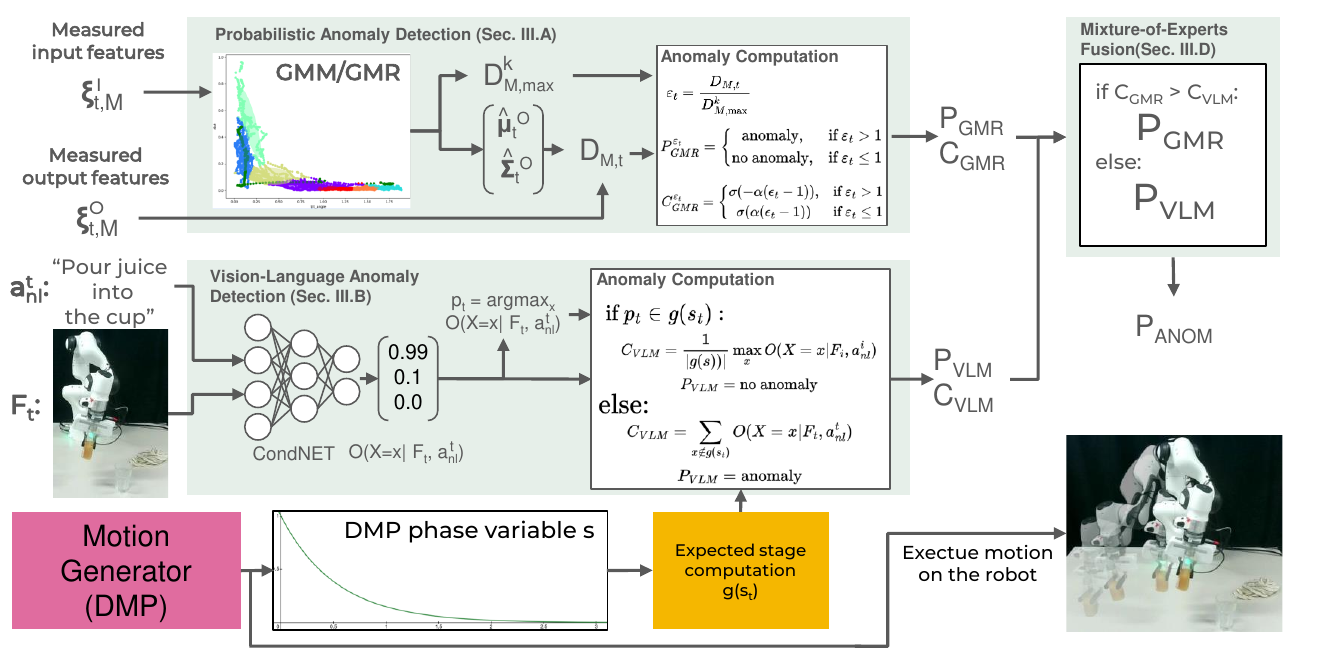}
    \caption{\textbf{System Architecture.}
    After training the VLM and GMR experts, we developed a mixture-of-experts anomaly detection pipeline that combines their strengths. For the GMR expert, we use measured input features to regress the predicted feature value $\hat{\mu}_t^O$ and covariance $\hat{\Sigma}_t^O$. We compute the maximum Mahalanobis distance for the mixture component associated with the observed input features and the Mahalanobis distance between observed and predicted output features. These values are combined to compute the anomaly prediction and confidence score, where $\sigma(x)$ is the sigmoid function and $\alpha$ controls the confidence magnitude. For the VLM, we leverage knowledge of the current action and camera observations. The VLM regresses the current motion state, which we compare to the expected state. Unlike prior work~\cite{sliwowski2024conditionnet}, we express the expected state as a function of action progress, enabling continuous motion policies instead of segmented primitives. Finally, we implement a late fusion strategy, weighted by prediction confidence, to merge both experts' outputs.}
    \label{fig:architecture}
\end{figure*}

In this work, we propose a mixture of experts that uses a late fusion strategy to merge the predictions from a probabilistic model (GMM) and a visual-language anomaly detector (VLM)~\cite{sliwowski2024conditionnet}. The VLM detects anomalies on a semantic level, e.g. a spill is when liquid is outside of the cup, while the GMM encodes the expected low-level dynamics of the task including process forces, and detects deviations from it. As a result, the individual detectors of our method are trained on different training data, which improves the performance of each detector.

\section{Methodology}
\subsection{Problem Formulation}
To avoid and react to errors during autonomous task execution, a robot must be able to identify unintended or faulty situations. Our proposed multimodal approach leverages proprioceptive and exteroceptive sensor measurements to achieve robust anomaly detection performance. Proprioception refers to contact force and end-effector pose collected with the robot, whereas exteroception includes visual features and object poses in the environment. Our proposed anomaly detection approach combines a probabilistic GMR-based detector $GMR_\theta(\bm{\xi}_i)$ with a visual-language classification-based action monitoring approach $VLM_\psi(F_i, a_{nl}^i)$. An overview of the mixture-of-experts anomaly detection pipeline can be seen in Figure~\ref{fig:architecture}.

Given a dataset $\mathcal{D} = \{\bm{d}_i\}_{i=1}^{N_{total}}$ of $N_{total}$ demonstrations, where each demonstration $\bm{d}_i$ consists of proprioceptive and exteroceptive sensory readings $\bm{p}_i$ and $\bm{e}_i$, a natural language task description $a_{nl}^i$, and the success status $success_i$, i.e., $\bm{d}_i = (\bm{p}_i, \bm{e}_i, a_{nl}^i, success_i)$, the goal is to learn two anomaly detection models, $GMR_\theta(\bm{\xi}_i)$ and $VLM_\psi(F_i, a_{nl}^i)$, which are fused into a mixture-of-experts model $MOE_{\theta, \psi}(\bm{o}_i)$, where $\bm{o}_i = (\bm{\xi}_i, F_i, a_{nl}^i)$ is the current observation during the execution, containing low-level features $\bm{\xi}_i$, the current video frame $F_i$, and the natural language task description. The parameter set $\theta$ includes the GMM's means $\bm{\mu}_k$, covariances $\bm{\Sigma}_k$, mixing coefficients $\pi_k$, and anomaly thresholds $\bm{D}^k_{M,max}$ for each of the $K$ components: $\theta = \{\bm{\mu}_k, \bm{\Sigma}_k, \pi_k, \bm{D}^k_{M,max}\}_{k=1}^K$. The set $\psi$ contains the weights and biases of the transformer-based vision-language anomaly detection model~\cite{sliwowski2024conditionnet}. To fuse the results from the GMR-based expert and the VLM-based expert, given $\bm{o}_i$, the mixture of experts compares the confidence score $C$ and anomaly prediction $P$ of both models, denoted with subscripts GMR and VLM.

To obtain the dataset $\mathcal{D}$, various demonstration techniques such as kinesthetic teaching~\cite{Lee2011}, haptic teleoperation~\cite{sliwowski2025reassemble}, and VR teleoperation~\cite{sliwowski2024conditionnet} can be used. Different experts in the mixture-of-experts model focus on different aspects of the data. For instance, since the GMM anomaly detection model relies on force and torque information, it is crucial that demonstrations capture accurate force and torque profiles. Thus, kinesthetic teaching is the most suitable technique. Conversely, the VLM anomaly detection model leverages only visual data, making it essential that the dataset accurately represents scene states during action execution. If the autonomous execution mode rarely includes a visible human, kinesthetic teaching becomes unsuitable, as it would always include the demonstrator. In such cases, VR or haptic teleoperation is preferable. Moreover, different experts may require varying amounts of data to model anomalies effectively. To ensure optimal training, we divide the dataset into two subsets: $\mathcal{D} = \mathcal{D}_{tele} \cup \mathcal{D}_{kin}$, where $\mathcal{D}_{tele} = \{d_i\}_{i=0}^{N_{tele}}$ is collected via teleoperation and $\mathcal{D}_{kin} = \{d_i\}_{i=0}^{N_{kin}}$ via kinesthetic teaching.

\subsection{Probabilistic anomaly detection}
For each skill, a probabilistic model of expected feature values is learned from kinesthetic user demonstrations. The feature values $\bm{\xi}$, including contact forces, relative distances, and orientations between the end-effector and objects during manipulation are computed from the proprioceptive and exteroceptive sensor readings collected during $N_{kin}$ task demonstrations of length $H_n$. The obtained training set containing the $L$ relevant features for a skill
\begin{equation}
\{\{\bm{\xi}_{h,n}\overunderset{H_n}{h=1}{\rbrace}\overunderset{N_{kin}}{n=1}{\rbrace} , \bm{\xi}_{h,n} \in \mathbb{R}^L 
\end{equation}
is encoded as a Gaussian Mixture Model (GMM), estimating the joint probability distribution of the training data set as a weighted sum of $K$ independent Gaussian components
\begin{equation}
\bm{\xi} \sim \sum_{k=1}^{K}\pi_k\mathcal{N}(\bm{\mu}_k, \bm{\Sigma}_k), 
\end{equation}
where $\pi_k, \bm{\mu}_k, \bm{\Sigma}_k$ are the mixing coefficient, feature mean, and covariance matrix of the $k$th Gaussian component. Similar to our previous work \cite{eiband2023collaborative, willibald2020collaborative}, we can decompose the features into an input and output set $\bm{\xi} = [\bm{\xi}^\mathrm{I}, \bm{\xi}^\mathrm{O}]^\mathrm{T}$, where the output features consist of the same sensor modality. We then use Gaussian Mixture Regression (GMR) at every time step $t$ during the task execution to infer the expected output feature vector $\mathbb{E}(\bm{\xi}_t^\mathrm{O}|\bm{\xi}_{t,\mathrm{M}}^\mathrm{I}) = \hat{\bm{\mu}}_t^\mathrm{O}$ along with the covariance matrix $\hat{\bm{\Sigma}}_t^\mathrm{O}$ given the measured input feature vector $\bm{\xi}_{t,\mathrm{M}}^\mathrm{I}$ by computing the conditional probability distribution $P(\bm{\xi}_t^\mathrm{O}|\bm{\xi}_{t,\mathrm{M}}^\mathrm{I}) = \mathcal{N}(\bm{\xi}_t^\mathrm{O}|\hat{\bm{\mu}}_t^\mathrm{O}, \hat{\bm{\Sigma}}_t^\mathrm{O})$. Using the measured output feature vector $\bm{\xi}_{t, \mathrm{M}}^\mathrm{O} $ at time $t$, we compute the Mahalanobis distance to quantify the deviation from the expected output feature vector with
\begin{equation}
    D_{M,t} = \sqrt{(\bm{\xi}_{t, \mathrm{M}}^\mathrm{O}-\hat{\bm{\mu}}_t^\mathrm{O})^\mathrm{T}(\hat{\bm{\Sigma}}_t^\mathrm{O})^{-1}(\bm{\xi}_{t, \mathrm{M}}^\mathrm{O}-\hat{\bm{\mu}}_t^\mathrm{O})}.
\end{equation}
Finally, to determine the anomaly prediction label $P_{GMR}$, we compute 
\begin{gather*}
    \epsilon_t = \frac{D_{M,t}}{D_{M,\mathrm{max}}^k}
\end{gather*}
 by dividing the Mahalanobis distance of the observed feature vector $D_{M,t}$ by the highest observed Mahalanobis distance $D_{M,\mathrm{max}}^k$ for the training data assigned to mixture component $k$ of the skill. We select $k$ as the mixture component maximizing $\argmax_k P(K = k| \bm{\xi}_{t,\mathrm{M}}^\mathrm{I})$ for the measured input feature vector. We set 
\begin{equation}
P_{GMR}^{\epsilon_t}= \left\{
\begin{array}{cl}
    \text{anomaly}, & \text{if } \epsilon_t > 1\\
    \text{no anomaly}, & \text{if } \epsilon_t \leq 1 \\
\end{array}\right.
\end{equation}
with prediction confidence
\begin{equation}
C_{GMR}^{\epsilon_t}= \left\{
\begin{array}{cl}
    \sigma(-\alpha(\epsilon_t-1)), & \text{if } \epsilon_t > 1\\
    \sigma(\alpha(\epsilon_t-1)) & \text{if } \epsilon_t \leq 1 \\
\end{array}\right.,
\end{equation}
where $\sigma(x)$ is the sigmoid function $\frac{1}{1+e^{x}}$. Hyperparameter $\alpha \in \mathbb{R}_{\geq0}$ can be chosen to scale the magnitude of the confidence score based on the number of observed demonstrations. We compute $\epsilon_t$ for all output sensor modalities and return $P_{GMR}=$ \textit{anomaly} if an anomaly is detected for at least one sensor modality. $C_{GMR}$ is set to the maximum confidence score across all anomaly predictions or the maximum confidence score if all predictions returned \textit{no anomaly} for all sensor modalities.

\subsection{Vision-Language anomaly detection}
Similarly, for each skill, we learn action preconditions and effects from the teleoperated dataset $\mathcal{D}_{tele}$ as described in~\cite{sliwowski2024conditionnet}. The input to the visual-language model is the current frame $F_i$, captured from an externally mounted camera, and the natural language description of the current skill, $a_{nl}^t$. Given the pair $(F_t, a_{nl}^t)$, the model outputs the probability distribution $O(X = x| F_t, a_{nl}^t)$ over three possible classes: \{pre, effect, unsatisfied\}. However, VLM predictions alone do not directly indicate whether an anomaly has occurred. For instance, if the model predicts {\tt pre} at the beginning of a motion, this is expected, whereas the same prediction at the end of the motion would signal an anomaly. In~\cite{sliwowski2024conditionnet}, this issue is addressed by segmenting skills into primitive motions and defining expected model predictions for each segment. However, for many complex or highly dynamic skills, segmenting into smaller units is challenging. Instead, in this work, we express the expected stage in terms of the motion phase.  

The phase of a motion can be represented in various ways depending on the chosen skill representation. For example,~\cite{path_variable} uses the distance along the path. In our work, we use Dynamical Movement Primitives (DMPs)~\cite{DMP} to model the motion policies, making the phase variable $s$ from the DMP a natural choice. We define the expected model predictions as a function of the phase variable, $g(s_t)$, where $g: s_t \longmapsto \{\text{pre}, \text{effect}, \text{unsatisfied}\}$, and manually specify it for each skill.  

To determine whether an anomaly has occurred, we compare the current model prediction, $p_i = \argmax_x O(X = x| F_t, a_{nl}^t)$, with the expected predictions given by $g(s_t)$. If $p_t$ belongs to the expected set, i.e., $p_t \in g(s_t)$, no anomaly is detected ($P_{VLM} = \text{no anomaly}$), and the anomaly prediction confidence is computed as  
\begin{equation}
    C_{VLM} = \frac{1}{|g(s_t)|} \max\limits_x O(X = x| F_t, a_{nl}^t),    
\end{equation}
where $\frac{1}{|g(s_t)|}$ serves to scale the confidence score. During experiments, we observed frequent fluctuations in model predictions during motion transition phases, such as when pouring liquid into a cup. By introducing $\frac{1}{|g(s_t)|}$, we can further reduce confidence in these periods, making the system rely more on other experts. If the current model prediction does not match the expected set, i.e., $p_t \notin g(s_t)$, an anomaly is detected ($P_{VLM} = \text{anomaly}$), and the anomaly confidence is computed as the sum of the probabilities assigned to the anomalous classes:  
\begin{equation}
    C_{VLM} = \sum\limits_{x \notin g(s_t)} O(X = x| F_t, a_{nl}^t).
\end{equation}

\subsection{Mixture-of-Experts}

Finally, we use a late fusion strategy to combine the predictions of both models. Since our mixture-of-experts (MoE) model consists of only two experts, we adopt a simple winner-takes-all approach, where the expert with the higher confidence determines the final prediction. Specifically, if the GMM anomaly prediction confidence is higher than that of the VLM, we take the GMM prediction as the final decision. Conversely, if the VLM confidence is higher or equal to that of the GMM, we use the VLM prediction:  

\begin{equation}
P = \left\{
\begin{array}{cl}
    P_{GMM} & \text{if } C_{GMM} > C_{VLM},\\
    P_{VLM} & \text{if } C_{VLM} \geq C_{GMM}. \\
\end{array}\right.
\end{equation}


\section{Experiments}
\subsection{Experimental setup}
We develop two distinct experimental scenarios to evaluate the performance of the proposed mixture-of-experts anomaly detection framework across vastly different tasks. Specifically, we consider an industrial and a household scenario.  

\begin{figure}[th]
  \centering
   \includegraphics[width=0.48\textwidth]{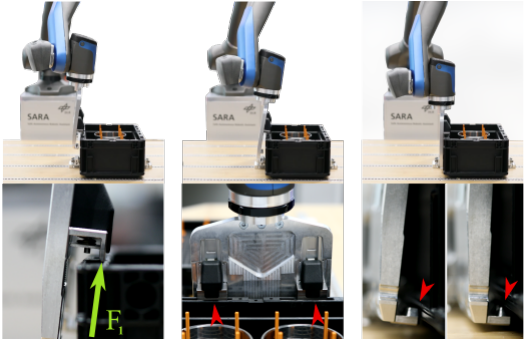}
  \caption{The different phases and challenges of the box grasping and locking task. As illustrated in the left column, the robot must apply force via the front part of the slides to move them up. It is important to maintain contact with the side wall of the box until the configuration in the middle is reached, without moving too close to the box as this blocks the sliding mechanism. After that, the locking pin must be pushed to compress the springs in the slides while rotating the gripper into the vertical configuration. The tight clearance between the box and the box locks on the lower part of the gripper will cause a collision when rotating too early.}
  \label{fig:Box_grasping_descr}
\end{figure}
Our industrial scenario in Fig.~\ref{fig:Box_grasping_descr} focuses on contact-rich manipulation, where the goal is to pick up a box using a specially designed gripper. The robot needs to precisely follow a sequence of motions to accurately grasp and lock the box within the passive gripper. The task consists of a sequence of four skills: (1) approaching the box, (2) pushing the gripper down along the side of the box to tension the springs inside the linear slides (two left columns in Fig.~\ref{fig:Box_grasping_descr}), (3) moving closer to lock onto the box, and (4) rotating into a vertical configuration to engage the gripper with the box. An overview of the different steps with a description of relevant aspects is shown in Figure~\ref{fig:Box_grasping_descr}. We define four possible anomalies in this scenario: (1) the gripper misses the box during the approach skill and fails to lock onto the box, (2) the linear slides of the gripper lock prematurely due to a hardware defect, (3) a user pushes the gripper during the free space motion, or (4) pulls it away while tensioning the springs of the gripper. In this scenario, we train the ConditionNET model on 107 successful and 71 unsuccessful skill examples. The GMM-based anomaly detection approach is trained on 3 kinesthetic demonstrations and 3 autonomous robot executions of the task, containing 4 skills each, which results in 24 successful skill examples. For each skill, the recorded 6D end-effector pose and the 3D contact force at the end-effector are encoded as a GMM with 2 mixture components. During execution, we condition the GMM on the measured end-effector pose to obtain the expected contact forces and compare them to the measured contact forces to detect unexpected deviations. To evaluate the performance of the different anomaly detection models, we then collect 47 successful and 35 unsuccessful skill executions, spanning the various anomaly cases.
\begin{figure*}[thpb]
    \centering
    \begin{subfigure}[b]{0.24\textwidth}
        \centering
        \includegraphics[width=\textwidth]{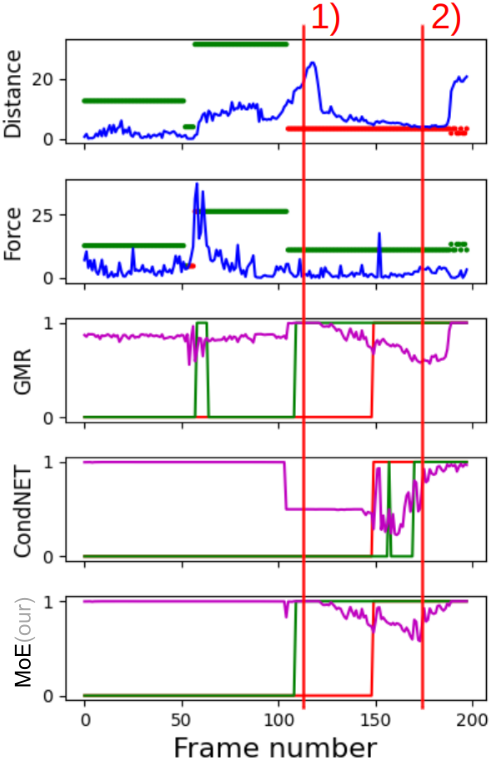}
        \caption{Overshoot}
        \label{fig:sfigure1}
    \end{subfigure}
    \hfill
    \begin{subfigure}[b]{0.24\textwidth}
        \centering
        \includegraphics[width=\textwidth]{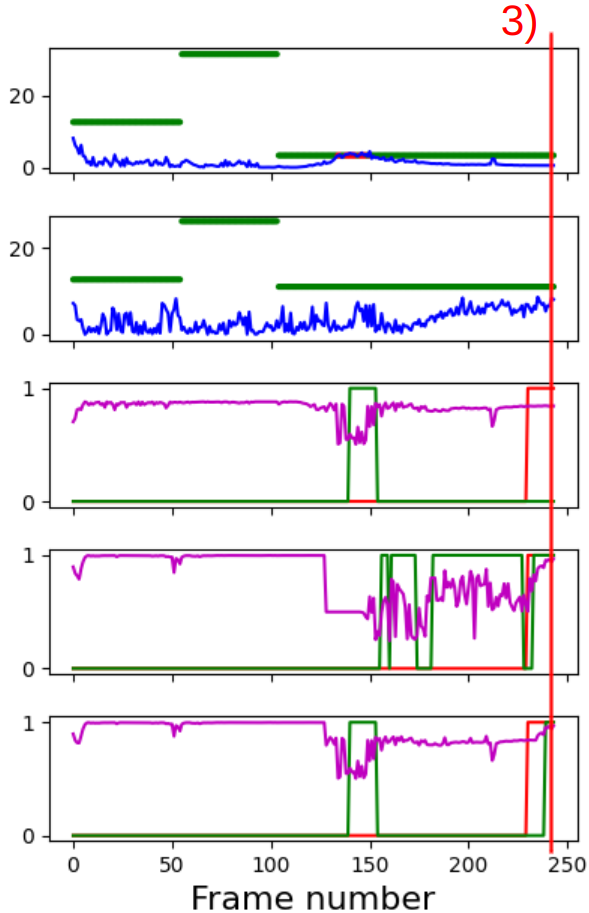}
        \caption{Drippring}
        \label{fig:sfigure2}
    \end{subfigure}
    \hfill
    \begin{subfigure}[b]{0.24\textwidth}
        \centering
        \includegraphics[width=\textwidth]{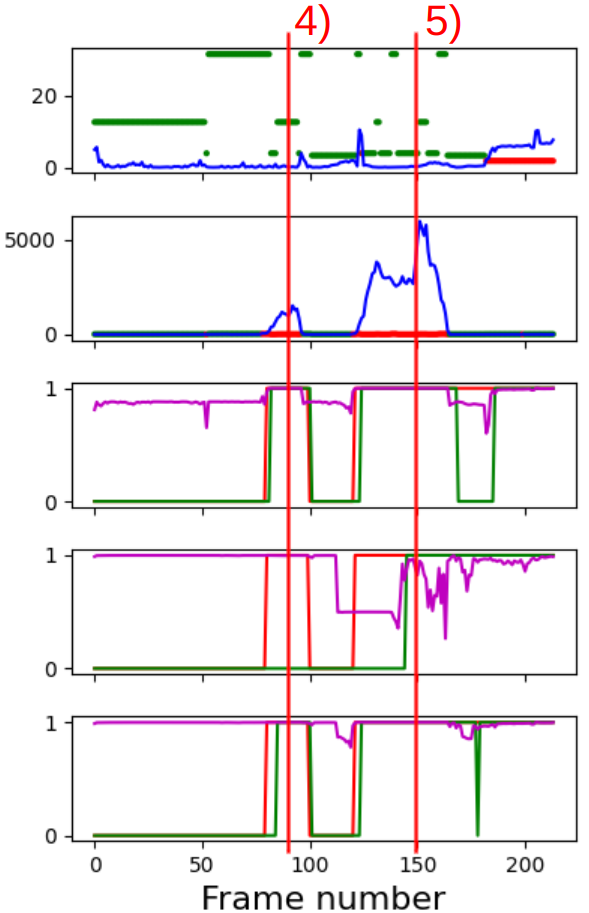}
        \caption{External perturbation}
        \label{fig:sfigure3}
    \end{subfigure}
    \hfill
    \begin{subfigure}[b]{0.24\textwidth}
        \centering
        \includegraphics[width=\textwidth]{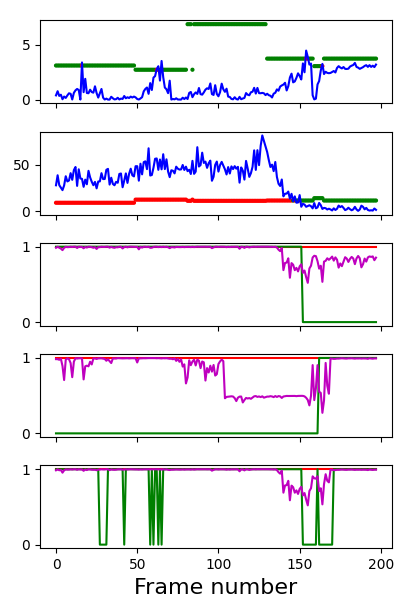}
        \caption{Empty bottle}
        \label{fig:sfigure4}
    \end{subfigure}
    \begin{subfigure}[b]{\textwidth}
        \centering
        \includegraphics[width=\textwidth]{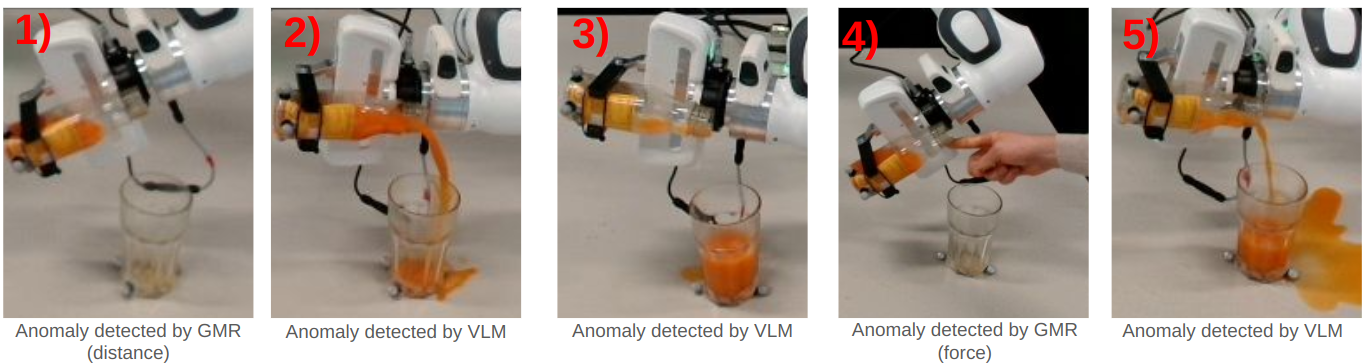}
        \caption{Anomaly detection frames of the GMR and VLM experts.}
        \label{fig:cases}
    \end{subfigure}
    \caption{Anomaly prediction results for different anomaly scenarios in the pouring task. The first two rows show the computed Mahalanobis distance in blue and the maximum Mahalanobis distance for the distance and force domain, respectively. If the Mahalanobis distance exceeds the threshold, The GMR-based detector triggers an anomaly.  The remaining three rows show the prediction confidence score in magenta and the anomaly prediction results for the respective methods GMR, ConditionNET, and our mixture-of-experts (MoE) in green. The ground truth anomaly result is indicated by the red line, where 0 means no anomaly and 1 means anomaly. Additionally,~\ref{fig:cases} shows snapshots from the executions when the anomalies are detected by the GMR and VLM experts. The respective time steps are marked with red vertical lines in the plots.}
    \label{fig:four_figures}
\end{figure*}

Our second scenario involves a juice-pouring task, where the robot's goal is to pour orange juice from a bottle into a glass. We consider two types of anomalies: (1) a spill occurs, and (2) the bottle is empty. Spills can result from various factors, such as the robot overshooting the cup, liquid dripping along the bottle’s edge during the final phase of pouring, or external disturbances caused by a human. In this scenario, we train the ConditionNET model on the (Im)PerfectPour dataset~\cite{sliwowski2024conditionnet}, which contains 406 successful and 138 unsuccessful task demonstrations. Additionally, we collect 4 successful task executions using kinesthetic teaching to train the GMM-based anomaly detection model on the recorded end-effector pose relative to the pouring target and the force on the end-effector using 10 mixture components. Finally, we collect 27 failed autonomous executions to evaluate the performance of the anomaly detection models.

\subsection{Quantitative Results}
\begin{table}[thpb]
\centering
\caption{Evaluation of the frame-wise detection performance of our mixture-of-experts anomaly detection approach (MoE) against the baselines GMR-based \cite{willibald2022multi}, and ConditionNET \cite{sliwowski2024conditionnet} for the box-grasping and the pouring task.}
\label{tab:total_anomaly_frame_evaluation}
\begin{tabular}{p{1.38cm}|llllll}
\toprule
 & \multicolumn{5}{c}{\textbf{Box-grasping}} \\
Method &  Acc &  Pre &  Rec &  F1 & F1@50 & Del \\
\midrule
MoE \color{gray}{(our)} & 88.1 & 96.6 & \textbf{82.6} & \textbf{88.3} & \textbf{86.4} & \textbf{0.47}  \\
GMR & \textbf{88.8} & \textbf{100} & 81.7 & 87.4 & 78.9 & 1.20 \\
CondNET & 79.8 & 95.9 & 73.2 & 81.6 & 75.0 & 1.23  \\
\midrule
 & \multicolumn{5}{c}{\textbf{Pouring}} \\
Method &  Acc &  Pre &  Rec &  F1 & F1@50 & Del \\
\midrule

MoE \color{gray}{(our)} & \textbf{88.7} & \textbf{88.7} & \textbf{88.1} & \textbf{87.2} & \textbf{84.7} & -0.3  \\
GMR & 84.5 & 86.9 & 81.0 & 83.3 & 76.7 & \textbf{-0.4} \\
CondNET & 75.8 & 88.0 & 67.2 & 70.2 & 69.3 & 0.4  \\

\bottomrule
\end{tabular}
\end{table}
We evaluate the frame-wise anomaly detection performance of our proposed MoE detector against the individual anomaly detection approaches integrated within the framework. To assess the performance, we report frame-wise accuracy, precision, recall, F1 score, F1 at 50\% overlap threshold, and detection delay in seconds averaged across all anomaly cases of a task. Detection delay is defined as the time interval between the first occurrence of an anomaly in the ground truth and the first time step at which the respective method triggers an anomaly. The F1 at 50~\% overlap threshold considers an anomaly detected if the frame-wise intersection over union (IoU) of anomaly ground truth and detection is larger than  50\%. To ensure confident anomaly classifications, we filter raw predictions of each method by taking the majority of predictions over a sliding time window of eight time steps.

Table~\ref{tab:total_anomaly_frame_evaluation} presents the quantitative evaluation results across various anomaly cases caused by diverse sources for the box-grasping and pouring task. In the box-grasping experiment, the MoE approach achieves frame-wise detection performance comparable to the GMR-based method while reducing detection delay by more than 60\% compared to both individual detectors. MoE also outperforms the other approaches in the F1@50\% score, indicating a more precise anomaly detection result. Notably, in the missed-box anomaly case, vision-based detection enables earlier anomaly identification compared to using only low-level features, whereas those are better suited to promptly detect force based anomalies (see Fig.~\ref{fig:box_grasping_time_anom}).

For the pouring task, MoE outperforms both other approaches for all frame-wise detection scores while maintaining a detection delay similar to GMR. Both MoE and GMR report a negative detection delay, which highlights the early detection capabilities of these methods, warning the system before the effect of an anomaly becomes visually apparent. In this scenario, these approaches can detect an imminent liquid spill if the bottle in the gripper is tilted outside the expected pouring region of the cup, allowing the system to intervene proactively.

\subsection{Qualitative Results}
The early detection of a spill by our MoE detector can be observed in Fig.~\ref{fig:sfigure1}, where the spill caused by overshooting the cup occurs at time step 150. However, since the robot leaves the expected pour region for the cup, while the bottle is already tilted (depicted in Fig.~\ref{fig:cases}-1)), the GMR-based detector in the upper row of Fig.~\ref{fig:sfigure1}, monitoring the relative end-effector position, triggers an anomaly shortly after time step 100. The higher prediction confidence score of the GMR-based detector outweighs the ConditionNET result, leading to anomaly detection within the MoE framework. Conversely, in the first half of the task, ConditionNET has a higher confidence score, outvoting a false positive prediction of the GMR-based detector. The dripping liquid failure case in Fig.~\ref{fig:sfigure2} occurs in the final pouring phase, as shown in Fig.~\ref{fig:cases}-3), when the bottle is nearly empty and the liquid sticks to the opening, flows along the bottleneck to then drip outside the cup. Since low-level sensor readings remain within the expected ranges, this anomaly is only visually detectable by ConditionNET, resulting in a true positive anomaly detection within the MoE framework. Around time step 150 during this task, a false positive anomaly detection is triggered, caused because both confidence scores of the experts are low and the measured distance at the beginning of the pouring crosses the boundary of the expected region. Fig.~\ref{fig:sfigure3} and Fig.~\ref{fig:cases}-4) depict a scenario, where the robot is pushed away from the cup in two different phases of the task, which is reliably detected by the GMR's force-based monitoring approach shown in the second row. After the second perturbation, when the robot continues with the normal task execution, GMR classifies the situation as normal. However, since a spill has already occurred during the second push, ConditionNET correctly identifies the anomaly in Fig.~\ref{fig:cases}-5), leading to an overall correct classification by the MoE framework. The final anomaly case for the pouring task is shown in Fig.~\ref{fig:sfigure4}, where the robot picks up an empty bottle. Since the expected force is not met during the first three quarters of the task, GMR continuously detects this deviation. Even though pouring with an empty bottle has not been present in the training data, ConditionNET detects that the effect of the task is not met when no liquid is poured an triggers an anomaly in the final phase of the task, leading to an overall improved result with the MoE.

\begin{figure}[thpb]
    \centering
    \begin{subfigure}[b]{0.48\linewidth}
        \centering
        \includegraphics[width=\textwidth]{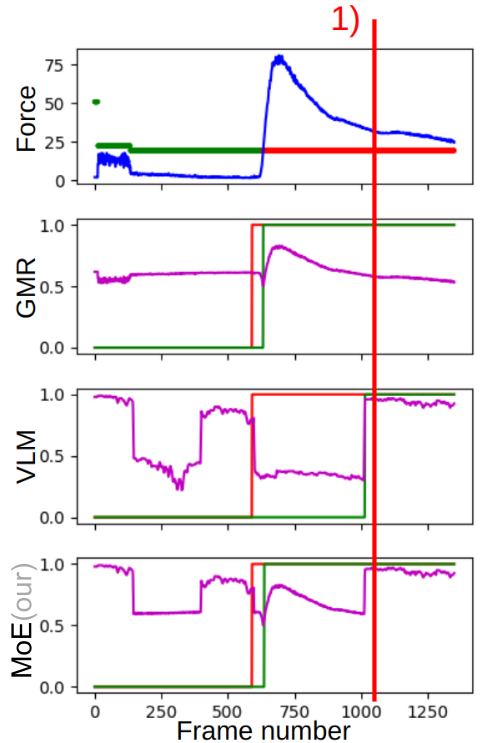}
        \caption{Slides locked}
        \label{fig:sfigure5}
    \end{subfigure}
    \hfill
    \begin{subfigure}[b]{0.48\linewidth}
        \centering
        \includegraphics[width=\textwidth]{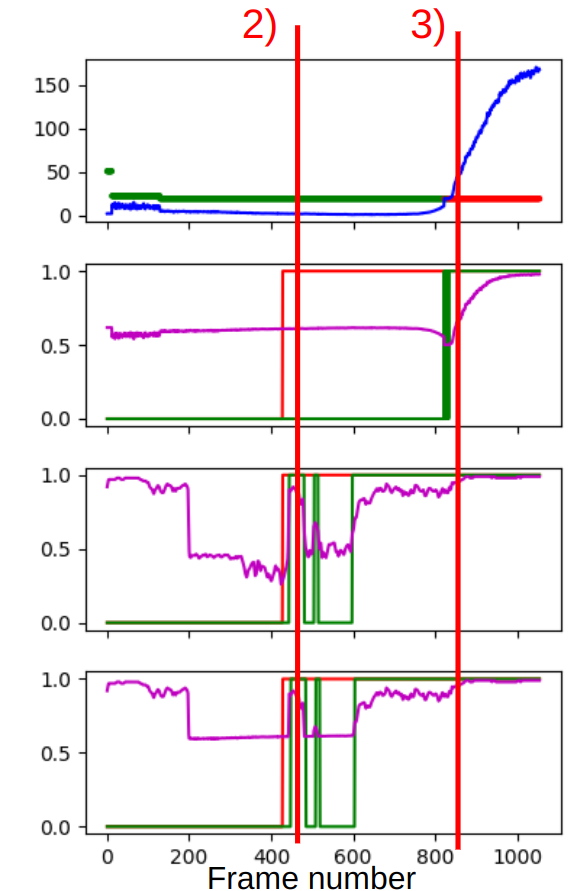}
        \caption{Box missed}
        \label{fig:sfigure6}
    \end{subfigure}
    \begin{subfigure}[b]{\linewidth}
        \centering
        \includegraphics[width=\textwidth]{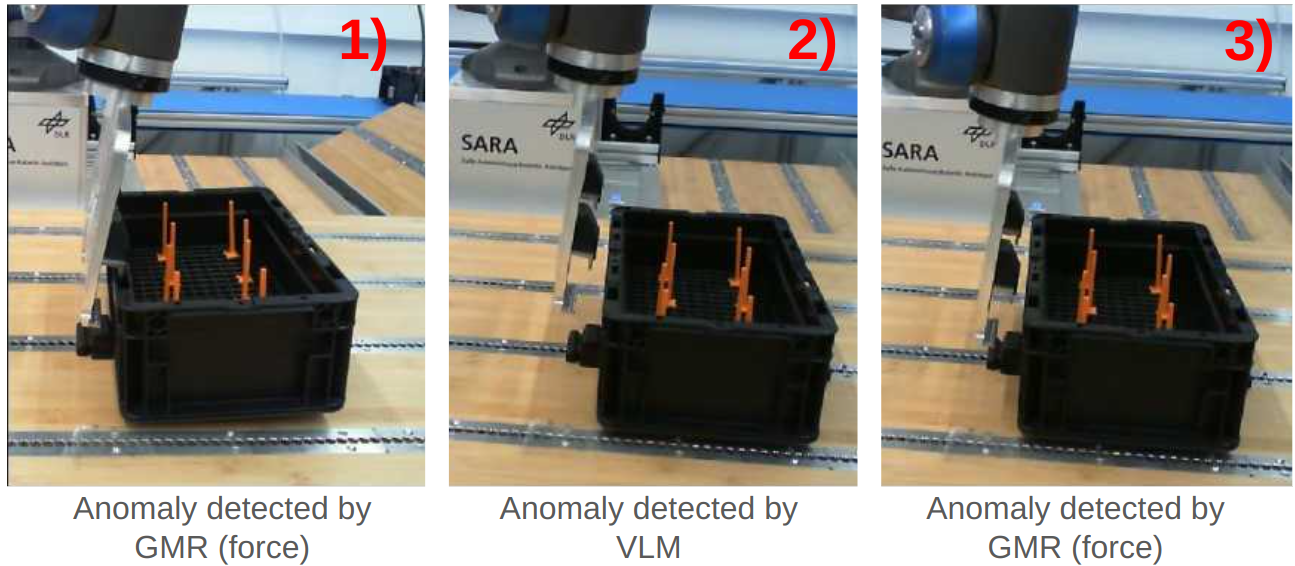}
        \caption{Anomaly detection frames of the GMR and VLM experts.}
        \label{fig:cases2}
    \end{subfigure}
    \caption{Anomaly prediction results for different anomaly scenarios in the industrial box grasping task. The upper row shows the computed Mahalanobis distance in blue and the maximum Mahalanobis distance for the force domain. If the Mahalanobis distance exceeds the threshold, The GMR-based detector triggers an anomaly. The remaining three rows show the prediction confidence score in magenta and the anomaly prediction results for the respective methods GMR, Condition NET, and our proposed mixture of experts (MoE) in green. The ground truth anomaly result is indicated by the red line, where 0 means no anomaly and 1 means anomaly. Additionally,~\ref{fig:cases2} shows snapshots from the executions when the anomalies are detected by the GMR and VLM experts. The respective time steps are marked with red vertical lines in the plots.}
    \label{fig:box_grasping_time_anom}
\end{figure}

Two representative anomaly situations for the box grasping task are depicted in Fig.~\ref{fig:box_grasping_time_anom}. In Fig.~\ref{fig:sfigure5}, a hardware defect blocks the gripper's slides from moving, causing an unexpected force reading that is correctly identified as anomalous by the GMR-based detector. Later in the task, ConditionNET also detects the anomaly when the push-down skill fails to reach its effect phase in the final quarter of execution (see Fig.~\ref{fig:cases2}-1). The combined MoE detection result significantly reduces the detection delay compared to ConditionNET alone. For the box missed case in Fig.~\ref{fig:sfigure6}, ConditionNET already detects that the gripper missed the box in the time step depicted in Fig.~\ref{fig:cases2}-2), when it should have made contact with the side wall. In contrast, the GMR-based approach only detects the anomaly later, once the force signal confirms the missed contact (Fig.~\ref{fig:cases2}-3). Again, the MoE framework substantially reduces the detection delay compared to the GMR-based detector alone.
\section{Conclusion}
We demonstrated the need for an approach that fuses anomaly predictions from two experts with different detection mechanisms on both an industrial and a household task. The GMR-based approach reliably identified anomalies by analyzing deviating contact forces, whereas the VLM-based detector registered even subtle visual deviations in the scene. In the experiments, we showed that the GMR- and VLM-based detectors complement each other in different scenarios, leading with our proposed MoE to a reduction of the detection delay of up to 60\% and an overall improved frame-wise detection performance. Future work could focus on integrating more modalities such as audio and depth perception, and combining further expert detectors with a modified voting system.
\bibliographystyle{IEEEtran}
\bibliography{main}
\end{document}